%% file: main.tex
\documentclass{article}

\usepackage{arxiv}

\usepackage[utf8]{inputenc} 
\usepackage[T1]{fontenc}    
\usepackage{hyperref}       
\usepackage{url}            
\usepackage{booktabs}       
\usepackage{amsfonts}       
\usepackage{nicefrac}       
\usepackage{microtype}      
\usepackage{lipsum}		
\usepackage{graphicx}
\usepackage{natbib}
\usepackage{doi}

\usepackage{amsmath,amssymb}
\usepackage{mathrsfs}
\usepackage{array}
\usepackage{textcomp}
\usepackage{stfloats}
\usepackage{url}
\usepackage{verbatim}
\usepackage{graphicx}
\usepackage{booktabs}
\usepackage{amsfonts}
\usepackage{makecell}
\usepackage{color}
\usepackage[figuresright]{rotating}
\usepackage{multirow}
\usepackage[normalem]{ulem}
\usepackage[caption=false,font=normalsize,labelfont=sf,textfont=sf]{subfig}
\newcommand{\tabincell}[2]{\begin{tabular}{@{}#1@{}}#2\end{tabular}}
\usepackage{caption}
\usepackage{subcaption}

\title{Artifact Feature Purification for Cross-domain Detection of AI-generated Images}

\author{
    Zheling Meng\\
    Center for Research on \\Intelligent Perception and Computing, \\
    State Key Laboratory of \\Multimodal Artificial Intelligence Systems, \\
    Institute of Automation, \\
    Chinese Academy of Sciences \\
    \texttt{zheling.meng@cripac.ia.ac.cn} \\
    \And
    Bo Peng \\
    Center for Research on \\Intelligent Perception and Computing, \\
    State Key Laboratory of \\Multimodal Artificial Intelligence Systems, \\
    Institute of Automation, \\
    Chinese Academy of Sciences \\
    \texttt{bo.peng@nlpr.ia.ac.cn} \\
    \And
    Jing Dong\thanks{Corresponding Author}  \\
    Center for Research on \\Intelligent Perception and Computing, \\
    State Key Laboratory of \\Multimodal Artificial Intelligence Systems, \\
    Institute of Automation, \\
    Chinese Academy of Sciences \\
    \texttt{jdong@nlpr.ia.ac.cn} \\
    \And
    Tieniu Tan \\
    Center for Research on \\Intelligent Perception and Computing, \\
    State Key Laboratory of \\Multimodal Artificial Intelligence Systems, \\
    Institute of Automation, \\
    Chinese Academy of Sciences \\
    \texttt{tnt@nlpr.ia.ac.cn} \\
}

\renewcommand{\shorttitle}{Artifact Feature Purification for Cross-domain Detection of AI-generated Images}

\hypersetup{
pdftitle={Artifact Feature Purification for Cross-domain Detection of AI-generated Images},
pdfauthor={Zheling Meng, Bo Peng, Jing Dong, Tieniu Tan},
pdfkeywords={AI-generated Image, Detection, Cross-domain},
}

\begin{document}
\maketitle
\vspace{1cm}
\input{0_abstract}
\input{1_intro}
\input{2_related_works}
\input{3_methods}
\input{4_experiments}

\bibliographystyle{unsrtnat}
\bibliography{main}  

\end{document}

%% file: 0_abstract.tex
\begin{abstract}
In the era of AIGC, the fast development of visual content generation technologies, such as diffusion models, bring potential security risks to our society. Existing generated image detection methods suffer from performance drop when faced with out-of-domain generators and image scenes. To relieve this problem, we propose Artifact Purification Network (APN) to facilitate the artifact extraction from generated images through the explicit and implicit purification processes. For the explicit one, a suspicious frequency-band proposal method and a spatial feature decomposition method are proposed to extract artifact-related features. For the implicit one, a training strategy based on mutual information estimation is proposed to further purify the artifact-related features. Experiments show that for cross-generator detection, the average accuracy of APN is $5.6\%\sim 16.4\%$ higher than the previous 10 methods on GenImage dataset and $1.7\%\sim 50.1\%$ on DiffusionForensics dataset. For cross-scene detection, APN maintains its high performance. Via visualization analysis, we find that the proposed method extracts flexible forgery patterns and condenses the forgery information diluted in irrelevant features. We also find that the artifact features APN focuses on across generators and scenes are global and diverse. The code will be available on GitHub.
\end{abstract}

\keywords{AI-generated Image, Detection, Cross-domain, Artifact, Purification}

%% file: 1_intro.tex
\section{Introduction}

\label{sec:intro}
As an important part of Artificial Intelligence Generated Content (AIGC), AI-generated images are drawing increasing attentions. The image generation methods, such as Generative Adversarial Networks (GANs) and Diffusion Models (DMs), endow us with powerful tools to generate high-fidelity images \citep{brock2018biggan, karras2019stylegan, song2020denoising, dhariwal2021diffusion, nichol2022glide, gu2022vector, rombach2022high, wukong, midjourney}. While they bring convenience and pleasure to our lives, the potential social harm cannot be underestimated. For example, some models may be used to maliciously generate fake news to create social panic \citep{sha2022fake} and some individuals generate biometrics of specific people without any permission \citep{mirsky2021creation, carlini2023extracting, zhu2023data}. This practical issue forces us to carry out researches on \emph{AI-generated image detection}. Recently, there have been many important developments on this topic. CNNSpot \citep{wang2020cnn}, F3Net \citep{qian2020thinking}, Patch5M \citep{ju2022fusing} and SIA \citep{sun2022information} are proposed successively to detect generated images by extracting synthesis features in the frequency or spatial modality. Faced with recently proposed high-quality generators such as DMs, methods like LNP \citep{liu2022detecting}, LGrad \citep{ma2023exposing} and SeDID \citep{ma2023exposing} are proposed to expose artifacts via transforming images into more intuitive representations. Inspired by multi-modal feature alignment technologies, the works \citep{amoroso2023parents, wu2023generalizable} propose the text-assisted detection methods. And the works \citep{chai2020makes, ricker2022towards, corvi2023detection} summarize the laws of synthesis traces left by common generators in the spatial and frequency modalities.

\begin{figure}[t]
  \centering
   \includegraphics[scale=0.17]{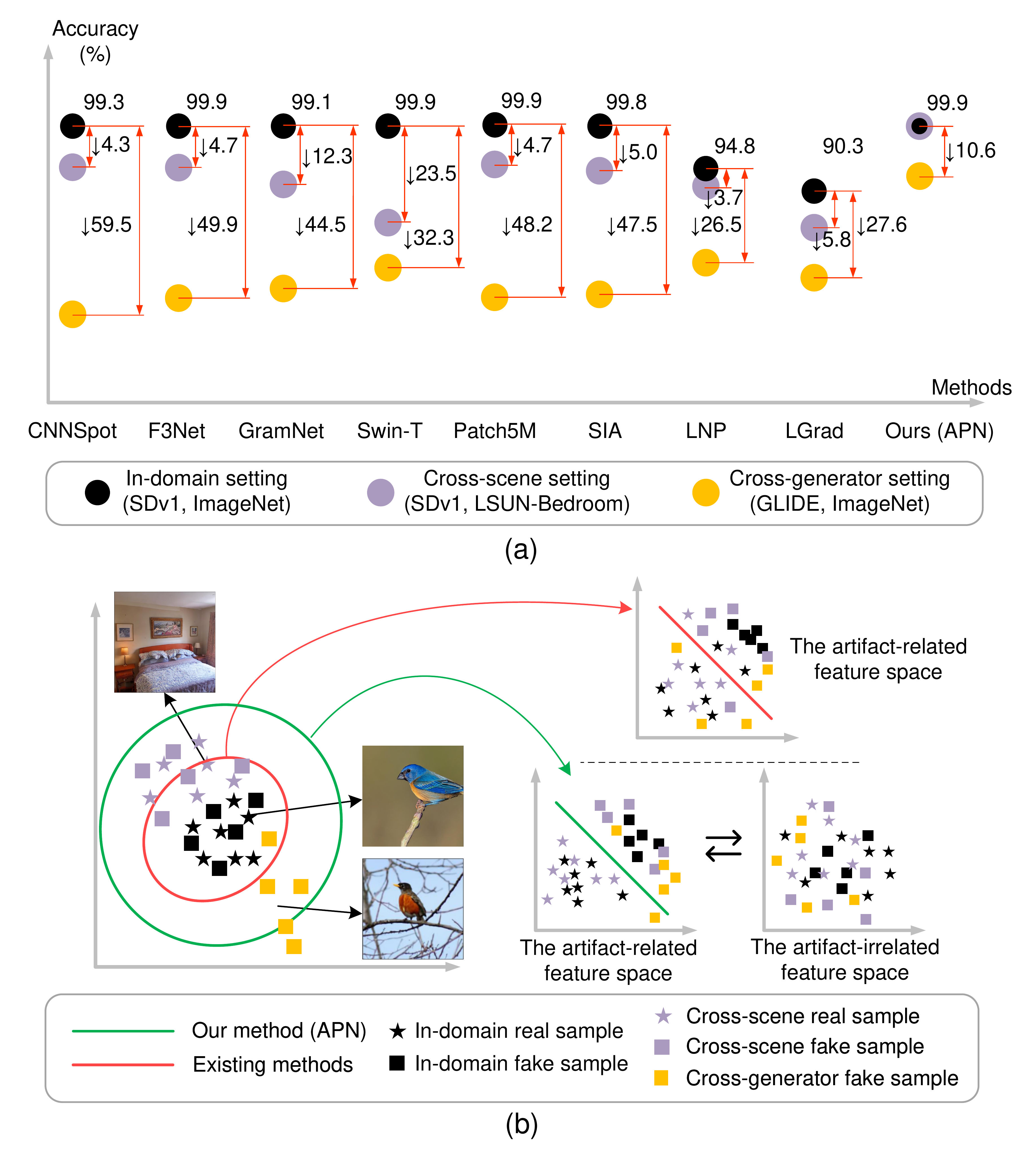}
   \caption{The cross-domain generated image detection. (a) The performance of CNNSpot \citep{wang2020cnn}, F3Net \citep{qian2020thinking}, GramNet \citep{liu2020global}, Swin-T \citep{liu2021swin}, Patch5M \citep{ju2022fusing}, SIA \citep{sun2022information}, LNP \citep{liu2022detecting}, LGrad \citep{tan2023learning} and ours (APN). They are trained on SD v1-generated samples under ImageNet scene \citep{deng2009imagenet}. Then they are tested on GLIDE-generated samples under ImageNet scene (cross-generator setting), and on SD v1-generated samples under LSUN-Bedroom scene \citep{yu2015lsun} (cross-scene setting). (b) The distribution difference between images of different domains, and the comparison between existing methods and ours. }
   \label{fig:intro_fig}
\end{figure}

In this paper, our focus is to improve cross-domain generated image detection performance. The domains here include not only the one formed by different generators (\emph{cross-generator detection}), but also the one formed by different image scenes (\emph{cross-scene detection}). Here, cross-scene detection refers to the ability of a method to detect the samples which have a different scene distribution compared with the training samples. Previous methods achieve excellent performance on in-domain samples. However, there is significant accuracy drop when they are tested on cross-domain samples. We show the problem by the experiments in Fig.\ref{fig:intro_fig}(a). Under the same scene, the detectors trained on SD v1-generated samples \citep{rombach2022high} cannot identify the GLIDE-generated samples \citep{nichol2022glide} with a similar accuracy. For the same generator, there is also an accuracy drop when they are trained on ImageNet scene \citep{deng2009imagenet} but tested on LSUN-Bedroom scene \citep{yu2015lsun}. The reason is that the feature spaces learned by existing methods are coupled with domain biased information. From the perspective of generators, they overfit to specific generator artifacts, which has been reported before \citep{ricker2022towards, lorenz2023detecting, corvi2023detection}. From the perspective of scenes, they overfit to object biased artifacts. The artifacts are coupled with the content features under in-domain scenes, leading to performance drop under other scenes. In practical application, it is impossible for the training of detectors to cover all generators and all scenes. More effective methods are needed to improve the ability to mine domain-agnostic forgery features.

In this paper, we propose \textbf{A}rtifact \textbf{P}urification \textbf{N}etwork, or APN, for cross-domain generated image detection. APN facilitates the artifact extraction through the explicit and implicit purification processes. \emph{The explicit purification} separates and extracts the artifact features from the frequency spectrum by the suspicious frequency-band proposal method (\emph{explicit purification in frequency}) and from the spatial features by the orthogonal decomposition method (\emph{explicit purification in space}). \emph{The implicit purification} help further aggregate the artifact information diluted in the features through a training strategy based on the mutual information theory. The purification processes improve the ability to acquire and analyze domain-agnostic artifact features, as shown in Fig.\ref{fig:intro_fig}(b). We detail the differences between previous methods and ours in Sec.\ref{sec: method discussion}. A series of experiments are performed in cross-domain settings. Based on two recently proposed datasets GenImage \citep{zhu2023genimage} and DiffusionForensics (DF) \citep{wang2023dire}, for the cross-generator setting, we train our network on one generator and test it on others under the same scene. For the cross-scene setting, we train our network on three generators under ImageNet scene and test it on the corresponding generators under LSUN-Bedroom scene. Results show that APN exhibits a better cross-scene and cross-generator detection performance compared with the previous 10 methods. For the cross-generator detection, the average accuracy of APN is $5.6\%\sim16.4\%$ higher than them on GenImage and $1.7\%\sim50.1\%$ on DF. For the cross-scene detection, the average accuracy of APN under out-of-scene samples is only 0.1\% lower than the one under in-scene samples. We also show and discuss the visualization results in Sec.\ref{sec:vis} to help understand \emph{the reason why APN can improve cross-domain detection performance} and \emph{the cross-domain commonalities of artifact features in space that APN focuses on}. Additional ablation experiments are conducted to analyze the roles of the key modules of APN. In summary, the main contributions in this paper are as follows: 
\begin{itemize}
    \item We point out that the performance drop of existing methods across generators and scenes is the main obstacle for their practical applications.
    \item We propose Artifact Purification Network. It extracts domain-agnostic artifact features by (1) Explicit Purification with the frequency-band proposals and the spatial feature decomposition, and (2) Implicit Purification with a training strategy based on the information theory. 
    \item The results show that compared with the previous 10 methods, APN achieves an improvement of average accuracy across generators on two datasets, and maintains the high performance on the cross-scene samples.
    
\end{itemize}

%% file: 2_related_works.tex
\section{Related Work}
\label{sec:related_works}

Existing methods usually regard generated image detection as a binary classification task. They can be summarized into two major categories, i.e. the single modality based and multiple modalities based methods.

\subsection{Detection with Single Modality}
Spatial image is the main modality considered in single modality based methods. Extracting features directly from images is an intuitive approach \citep{wang2020cnn, chai2020makes, touvron2021training} and their performance is usually unsatisfactory. Then more specialized methods are proposed. RFM \citep{wang2021representative} is proposed to enlarge its attention regions. Ensemble learning \citep{mandelli2022detecting} provides multiple orthogonal views of an image to enrich synthetic features. The self-information \citep{sun2022information} is used to generate attention weights to capture more critical forgery cues. ADAL \citep{li2022artifacts} is proposed to disentangle and reconstruct artifacts and enhance training by adversarial learning. And the work \citep{hua2023learning} explores the interpretability by associating spatial patches and feature channels. There are also many works based on preprocessed spatial images. Luo et al. \citep{luo2021generalizing}, Fei et al. \citep{fei2022learning} and Xi et al. \citep{xi2023ai} introduce SRM filter \citep{fridrich2012rich} to separate the high-frequency noise from a forgery image. Liu et al. \citep{liu2022detecting} observe that the noise pattern of real images exhibits similar patterns and propose LNP for detection. Guo et al. \citep{guo2023data} propose a data augmentation framework to mine structured features of forgery faces. DIRE \citep{wang2023dire} and SeDID \citep{ma2023exposing} expose the differences between real and fake images via a pre-trained diffusion model. Different from them, LGrad \citep{tan2023learning} represents the artifacts produced by generation models as the gradients.

Frequency spectrum is another modality considered in single modality based detection methods. Previous studies \citep{ricker2022towards, corvi2023intriguing, corvi2023detection} have demonstrated that generators leave distinguishable artifacts in frequency. Some works \citep{frank2020leveraging, li2021frequency} dedicate to learn frequency differences between real and fake images directly. F3Net \citep{qian2020thinking} divides the spectrum into high, middle and low frequency bands, and analyzes them separately. The work \citep{jia2022exploring} proposes a frequency adversarial attack method to improve detection performance. However, these works are limited to face forgery detection and there is still a lack of explorations to detect generation for more diverse scenes and more recent generators.

\subsection{Detection with Multiple Modalities}
Based on the researches above, a natural idea is to combine spatial image and frequency spectrum for the detection. RGB-frequency attention module \citep{chen2021local} fuses features in both spatial and frequency modalities to collaboratively learn comprehensive
representation. Li et al. \citep{li2021frequency} propose an adaptive frequency information mining block to provide frequency clues for spatial feature extraction. Tian et al. \citep{tian2023frequency} extract consistent representations between augmented samples from both modalities. Gu et al. \citep{gu2022exploiting} propose PatchDCT to fuse local frequency features together with spatial images. The work \citep{woo2022add} proposes to conquer high frequency information loss of low quality images using knowledge distillation. Some researches try to introduce information of more modalities to improve the detection generalization. For example, LSD \citep{wu2023generalizable} is proposed to augment the training images with textual labels under the guidance of languages and formulate the classification problem as an identification one. Amoroso et al. \citep{amoroso2023parents} suggest to analyze the semantics of textual descriptions and low-level perceptual cues of fake images. Yin et al. \citep{yin2023dynamic} investigate spatio-temporal inconsistency caused by motion disturbance in deepfake videos. Cai et al. \citep{cai2023glitch} contribute a audio–visual forgery detection and localization benchmark. And a cross-modality knowledge distillation method is proposed in \citep{wang2022forgerynir} for face forgery detection in the near-infrared modality.

%% file: 3_methods.tex
\section{Methods}

\subsection{Overview}

\begin{figure*}[t]
  \centering
   \includegraphics[scale=0.21]{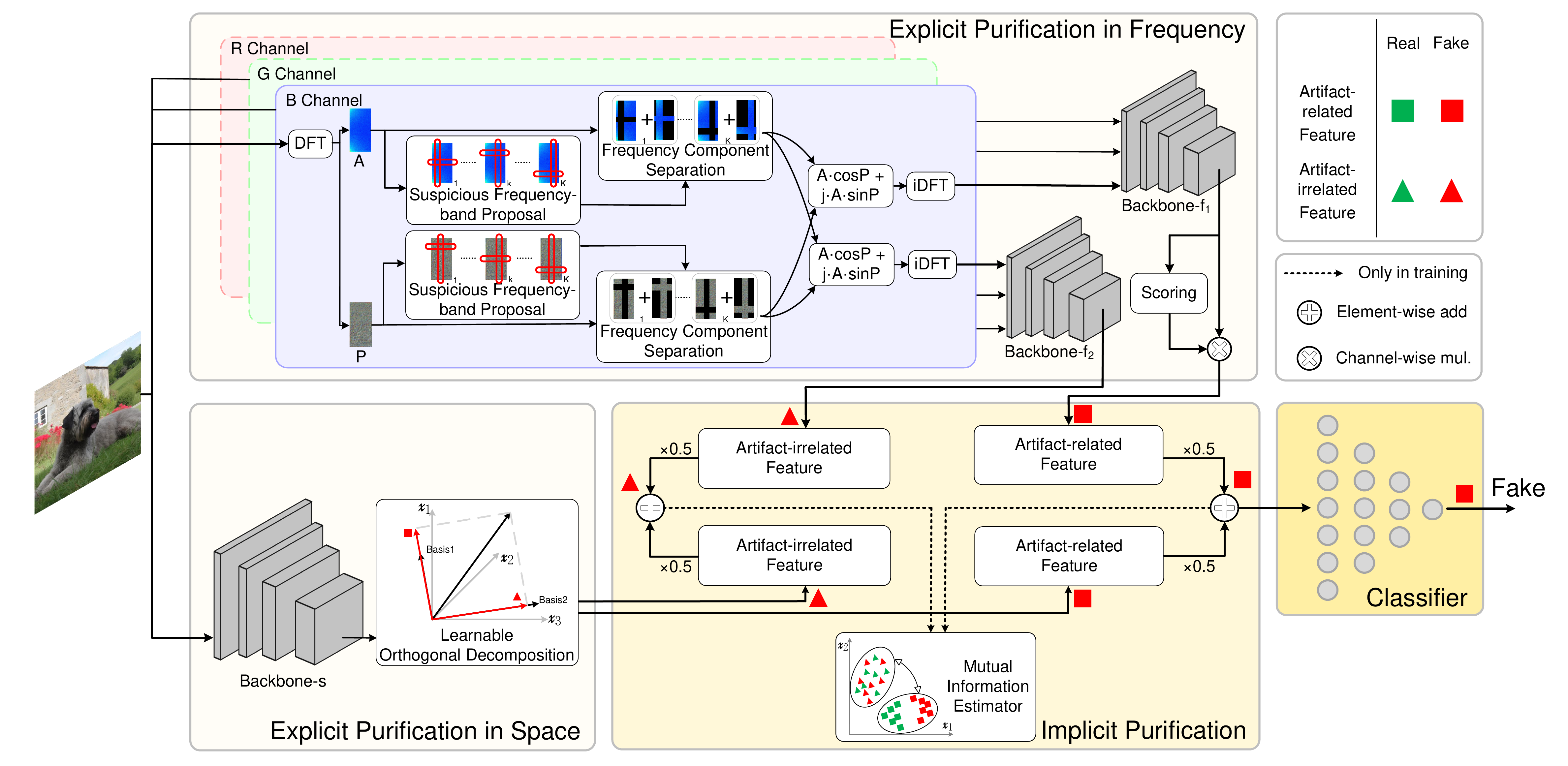}
   \caption{The framework of Artifact Purification Network (APN). APN facilitates the artifact extraction process through the explicit and the implicit purification. The explicit one separates artifacts and extracts the features from frequency and spatial modalities, where a suspicious frequency-band proposal method and a learnable spatial feature decomposition method are proposed for them respectively. The implicit one further purifies the artifact-related features based on the mutual information estimation. }
   \label{fig:framework}
\end{figure*}

The framework of our proposed APN is shown in Fig.\ref{fig:framework}. APN includes three parts, i.e. the explicit purification, the implicit purification and a classifier. The explicit purification is divided into two branches, separating the artifact features from frequency spectra and spatial images respectively. The implicit purification aligns the corresponding features from both branches and further purifies the artifact features by a mutual information estimation based training strategy. An important difference between the implicit and explicit purification is that the former is achieved by designing a training strategy and is discarded during the inference stage. Finally, the averaged artifact-related features are used to output a decision probability. 

\subsection{Explicit Purification}
\label{sec:explicit purification}

\subsubsection{Explicit Purification in Frequency}
\label{sec:freq branch}
In this branch, a suspicious frequency-band proposal method is proposed to separate the frequency components containing artifacts. Before that, Discrete Fourier Transform (DFT) is applied to transform inputs from spatial to frequency modality. Given an input image $I$ with $I_R$, $I_G$ and $I_B$ three channels, the spectrum of its certain channel $c$ is $\mathcal{F}_{c}$ where both $I_c$ and $\mathcal{F}_{c}$ have width $M$ and height $N$. Considering that $I_c$ only contains real values, $\mathcal{F}_{c}$ is Hermitian-symmetric. It means that the negative frequency terms are just the complex conjugates of the corresponding positive frequency terms, and the negative-frequency terms are therefore redundant. It should be noted that it cannot hold true on both $u$ and $v$ axes at the same time. Taking this advantage, we only need to calculate $\mathcal{F}_{c}(u, v)$ within the range of $0\leq u \leq \lfloor \frac{M}{2} \rfloor$ and $0\leq v \leq N - 1$. More importantly, it can ensure that inverse DFT of the spectrum after frequency separation also only contains real values. Taking into account the truth that amplitude and phase have distinct meanings and characteristics in signal processing, we calculate the amplitude $A_c$ and phase $P_c$ of $\mathcal{F}_{c}$ respectively.

APN will give the positions of the $K$ frequency-band proposals for $A_c$ and $P_c$ along $u$ and $v$ axes respectively. Take $A_c$ alone $u$ axis as an example. We use a batchnorm and several residual convolutions to obtain the feature map $\mathcal{M}_{A_c}$ of $A_c$. Define One-sided Feature Pooling $OFP(\cdot)$ as a convolution with a kernel size $(3 \times N)$ and a padding size $(1 \times 0)$. $OFP(\cdot)$, along with a fully connected layer and a Sigmoid activation, is applied on $\mathcal{M}_{A_c}$ to predict the normalized starting point $PPS_{A_c,u}^{i1}$ and ending point $PPS_{A_c,u}^{i2}$ of the proposal $PPS_{A_c,u}^{i}$, where $i=0,1,...,K-1$ denotes $i$-th proposal. In order to allow different proposals to focus on different frequency bands, we further assign each proposal a prior frequency position. 
It is $i$-th equal-division point in the range $[0,1]$ for $i$-th proposal. By modifying the kernel size of $OFP(\cdot)$ as ($\lfloor \frac{M}{2} \rfloor \times 3$) and the padding size as $(0 \times 1)$, we can get $PPS_{A_c,v}^{i}$ along $v$ axis. Similarly, $PPS_{P_c,u}^{i}$ and $PPS_{P_c,v}^{i}$ can also be obtained.

The artifact components will be separated from $A_c$ and $P_c$ according to the proposals. Specifically, the $i$-th artifact-\textbf{r}elated components $A_c^{i,r}$ and $P_c^{i,r}$, and the $i$-th artifact-\textbf{ir}related components $A_c^{i,ir}$ and $P_c^{i,ir}$ are as follows:
\begin{equation}
\label{eq:seperate}
    \begin{split}
        A_c^{i,r} = \mathbb{I}_{A_c}^{i} * A_c \qquad&
        P_c^{i,r} = \mathbb{I}_{P_c}^{i} * P_c\\
        A_c^{i,ir} = (1 - \mathbb{I}_{A_c}^{i}) * A_c \qquad&
        P_c^{i,ir} = (1 - \mathbb{I}_{P_c}^{i}) * P_c\\
    \end{split}
\end{equation}
where $*$ is the element-wise multiplication, and $\mathbb{I}_{A_c}^{i}$ and $\mathbb{I}_{P_c}^{i}$ are the cross shaped masks for $A_c$ and $P_c$ respectively, as seen in Fig.\ref{fig:framework}. They are generated according to the predicted starting and ending points of the proposals:
\begin{equation}
    \mathbb{I}_{A_c}^{i} (u, v) = \begin{cases}
            1 \qquad &if \quad PPS_{A_c,u}^{i1} \leq u / \lfloor \frac{M}{2} \rfloor \leq PPS_{A_c,u}^{i2} \\
            1 \qquad &if \quad PPS_{A_c,v}^{i1} \leq v / (N - 1) \leq PPS_{A_c,v}^{i2} \\
            0 \qquad &otherwise
    \end{cases}
\end{equation}
\begin{equation}
    \mathbb{I}_{P_c}^{i} (u, v) = \begin{cases}
            1 \qquad &if \quad PPS_{P_c,u}^{i1} \leq u / \lfloor \frac{M}{2} \rfloor \leq PPS_{P_c,u}^{i2} \\
            1 \qquad &if \quad PPS_{P_c,v}^{i1} \leq v / (N - 1) \leq PPS_{P_c,v}^{i2} \\
            0 \qquad &otherwise.
    \end{cases}
\end{equation}
Then we reassemble the components and transform them back to the spatial modality through inverse DFT ($iDFT(\cdot)$):
\begin{equation}
\label{eq:reassemble}
    \begin{split}
        I_c^{i,r} &= iDFT(A_c^{i,r}\cdot \sin{P_c^{i,r}} + j\cdot A_c^{i,r}\cdot \cos{P_c^{i,r}})\\
        I_c^{i,ir} &= iDFT(A_c^{i,ir}\cdot \sin{P_c^{i,ir}} + j\cdot A_c^{i,ir}\cdot \cos{P_c^{i,ir}})\\
    \end{split}
\end{equation}
where $j$ denotes the imaginary unit. By Eq.\ref{eq:reassemble}, we get the images $I_c^{i,r}$ with suspicious frequency components and the images $I_c^{i,ir}$ without them, where $i=0,1,...,K-1$. $I_c^{i,r}$ and $I_c^{i,ir}$ for $R$, $G$ and $B$ channels are concatenated along the channel dimension into $I^{i,r}$ and $I^{i,ir}$ respectively. Both $I^{i,r}$ and $I^{i,ir}$ are three-channel images. Two backbones are used to extract the features of them to obtain the feature vectors of $I^{i,r}$ and $I^{i,ir}$. 

To assess the importance of the suspicious artifacts in different proposals, the confidence scores for the feature vectors of $I^{i,r}$ are predicted. Specifically, a multi-head self-attention layer \citep{Vaswani2017attention} is used first, where the inputs $q$, $k$ and $v$ are all the concatenation of the feature vectors of $I^{i,r}$. Then its outputs are fed into three fully connected layers with a Sigmoid activation at the end to predict their confidence scores. The artifact-related feature vector $f^{r}_f$ of the input $I$ in frequency is the weighted average of the feature vectors of $I^{i,r} (i=0,1,...,K-1)$ with the corresponding scores. And the artifact-irrelated feature vector $f^{ir}_f$ of the input $I$ in frequency is the average of the feature vectors of $I^{i,ir} (i=0,1,...,K-1)$.

\subsubsection{Explicit Purification in Space}
\label{sec:spatial branch}
The key to achieving cross-domain generalized detection is the extraction of cross-domain generalized features. The spatial modality contains rich artifacts but they are often coupled with image contents. Previous works like LNP \citep{liu2022detecting} and LGrad  \citep{tan2023learning} expose the artifacts from images using pre-processing tools such as a denoising network. We notice that artifact-related and artifact-irrelated features should not overlap with each other ideally, as the name indicates. It implies an orthogonal relationship in the feature space. In this subsection, we propose a learnable feature orthogonal decomposition method. Given an input image $I$, this branch first extracts its feature vector $f_s \in \mathcal{R}^{C \times 1}$ by a backbone where $C$ is the dimension. Then $f_s$ is decomposed as the artifact-related feature $f^{r}_s \in \mathcal{R}^{C \times 1}$ and the artifact-irrelated feature $f^{ir}_s \in \mathcal{R}^{C \times 1}$. Specifically, there are two learnable mutually orthogonal sets of vector bases $\Omega^{r}=\{\omega^{r}_1, \omega^{r}_2,...,\omega^{r}_{m_r}\}$ and $\Omega^{ir}=\{\omega^{ir}_1, \omega^{ir}_2,...,\omega^{ir}_{m_{ir}}\}$, where $\omega_j^{r} \in \mathcal{R}^{C \times 1}$ $ (j=1,2,...,m_r)$, $\omega_k^{ir} \in \mathcal{R}^{C \times 1}$ $(k=1,2,...,m_{ir})$, $\omega^{r}_j \cdot \omega^{ir}_k = 0$ $(j=1,2,...,m_r, k=1,2,...,m_{ir})$ and $m_r + m_{ir} = C$. Then the process of decomposing can be expressed as follows:
\begin{equation}
    \begin{split}
        f^{r}_s &= \sum_{j=1}^{m_r}(\omega^{r}_j)^Tf_s\cdot\omega^{r}_j \\
        f^{ir}_s &=\sum_{k=1}^{m_{ir}}(\omega^{ir}_k)^Tf_s\cdot\omega^{ir}_k.
    \end{split}
\end{equation}

\subsection{Implicit Purification}
\label{sec:implicit purification}
In \ref{sec:explicit purification}, both branches extract artifact-related and artifact-irrelated features. We have the following two insights for them. On the one hand, a spatial image and its spectrum have a mapping relationship established on DFT. The frequency features with (without) the artifact components should be consistent with the spatial features with (without) the artifact components. The corresponding features from both branches should have unified representations. In turn, the unification can also help the branches learn to separate features from the two modalities better in the training stage. On the other hand, the explicit purification cannot ensure that the artifact information diluted in the features is also purified well (seen in \ref{sec:vis}). The artifact-related features may be still domain-biased and the artifact-irrelated features may still contain artifact information. Therefore, we propose the implicit purification to align and further purify the features.

To unify the feature representations, we require that the corresponding features from both branches are consistent. Specifically, the loss $\mathcal{L}_{align}$ is employed here, which has the form of Mean Squared Error:
\begin{equation}
\label{eq:pull}
    \mathcal{L}_{align} = \frac{1}{C|\mathcal{D}|}\sum_{\mathcal{D}}||f_f^r - f_s^r||^2_2 + \frac{1}{C|\mathcal{D}|}\sum_{\mathcal{D}}||f_f^{ir} - f_s^{ir}||^2_2 
\end{equation}
where $\mathcal{D}$ is a training batch with the size $|\mathcal{D}|$ and $C$ is the dimension of the feature space.

To purify the artifact information in the features, we calculate the mutual information between the artifact-related and artifact-irrelated features, and then minimize it. Usually, minimizing mutual information is transformed into minimizing its upper bound. Following \citep{cheng2020club}, the loss to minimize the variational upper bound of the mutual information between $f^r$ and $f^{ir}$ is: 
\begin{equation}
\label{eq:push}
    \begin{split}
    \mathcal{L}_{mi} = \frac{1}{|\mathcal{D}|}\sum_{\{f^r, f^{ir}\} \in \mathcal{D}}  \Big[\log  q_\theta(f^r|f^{ir})  
    -\frac{1}{|\mathcal{D}|}\sum_{\{f^{r'}\} \in \mathcal{D}}  \log q_\theta(f^{r'}|f^{ir})  \Big]
\end{split}
\end{equation}
where $q_\theta(f^r|f^{ir})$ is a neural network with parameters $\theta$ to approximate the real distribution $p(f^r|f^{ir})$, and $f^{r}$ and $f^{ir}$ are the average of the corresponding vectors from the two branches. The loss of training $q_\theta(f^r|f^{ir})$ to ensure the validity of the estimation is:
\begin{equation}
\label{eq:mie}
    \mathcal{L}_{q} = \frac{1}{|\mathcal{D}|}\sum_{\{f^r, f^{ir}\} \in \mathcal{D}} -\log q_\theta(f^r|f^{ir}).
\end{equation}

\subsection{Training Loss}
To implement our proposed method, two networks need to be trained, i.e. the proposed APN and the estimator $q_\theta$. 

The training loss of APN contains five parts, $\mathcal{L}_{ce}$, $\mathcal{L}_{band}$, $\mathcal{L}_{basis}$, $\mathcal{L}_{align}$ and $\mathcal{L}_{mi}$:
\begin{equation}
\label{eq:loss}
    \mathcal{L}_{APN}= \mathcal{L}_{ce} + \mathcal{L}_{band} + \mathcal{L}_{basis} + \mathcal{L}_{align} + \mathcal{L}_{mi}.
\end{equation}
Among them, $\mathcal{L}_{ce}$ is the cross entropy loss for classification, and $\mathcal{L}_{band}$ is a loss proposed to ensure the validity of the frequency band intervals of the proposals:
\begin{equation}
\label{eq:band}
\begin{split}
    \mathcal{L}_{band} = &\frac{1}{|\mathcal{D}|}\sum_{\mathcal{D}}\sum_{c\in\{R,G,B\}}\sum_{I\in\{A,P\}}\sum_{a\in\{u,v\}}\sum_{i=0}^{K-1} \\
    & \max\left(- (PPS_{I_{c},a}^{i2}- PPS_{I_{c},a}^{i1} - \epsilon), 0\right)
\end{split}
\end{equation}
where $\epsilon$ is a small positive real number. $\mathcal{L}_{basis}$ is proposed to ensure the orthogonality of $\Omega^r$ and $\Omega^{ir}$ in \ref{sec:spatial branch}:
\begin{equation}
    \mathcal{L}_{basis} = \frac{1}{m_r m_{ir}}\sum_{i=1}^{m_{r}}\sum_{j=1}^{m_{ir}} |(\omega^{r}_i)^T\omega^{ir}_j|.
\end{equation}
And $\mathcal{L}_{align}$ and $\mathcal{L}_{mi}$ are described as Eq.\ref{eq:pull} and Eq.\ref{eq:push} respectively. The training loss of $q_\theta$ is Eq.\ref{eq:mie}.

\subsection{Discussion}
\label{sec: method discussion}
There are three main innovations in APN, i.e. the frequency-band proposal method, the spatial feature orthogonal decomposition method, and the implicit purification based on the mutual information estimation. They have obvious differences in motivations and implementations compared with previous methods. F3Net \citep{qian2020thinking} divides the frequency into three sections to analyze the forgery patterns in the sections respectively. In contrast, APN proposes suspicious frequency bands containing potential artifacts according to inputs, and separates them from images to form forgery features. LNP \citep{liu2022detecting} simply concatenates spectral amplitude and phase from noise residuals, while APN is dedicated to discovering, separating and reassembling suspicious artifacts from amplitude and phase respectively. LNP \citep{liu2022detecting} and LGrad \citep{tan2023learning} expose artifacts from images using pre-trained models, while APN achieves it based on the relationship between artifact-relevant and -irrelevant features. SIA \citep{sun2022information} highlights the informative features by computing self-information, while APN estimates mutual information between decoupled features to further separate domain-agnostic artifact information. Last but not least, APN enhances the generalization of relevant features with the help of irrelevant features, while other methods study how to directly extract generalized features. In the experiments, we will see that the innovations make the cross-domain performance of APN better than previous methods significantly.

%% file: 4_experiments.tex
\section{Experiments}
\label{sec:experiments}
\subsection{Experiment Settings}
\subsubsection{Datasets}
\textbf{GenImage} \citep{zhu2023genimage} is a large-scale generated image dataset. It contains 1,331,167 real images from ImageNet \citep{deng2009imagenet} and 1,350,000 fake images generated by 8 generators using ImageNet classes. GenImage is divided into 8 non-overlapping subsets with a similar number of images. Each subset contains fake images from one of generators and is divided into a training and evaluating set further. \textbf{DiffusionForensics (DF)} \citep{wang2023dire} is another used dataset. For training, it provides 40,000 real images from LSUN-Bedroom \citep{yu2015lsun} and 40,000 images generated by ADM \citep{dhariwal2021diffusion}. For evaluating, it provides the fake images generated by 8 generators. Each generator generates 1,000 images under LSUN-Bedroom scene, except that the number of images generated by Midjourney \citep{midjourney} is 100. And 1,000 real images are also included.

\subsubsection{Baselines}
10 methods are used as the compared methods, i.e. Spec \citep{zhang2019detecting}, CNNSpot \citep{wang2020cnn}, F3Net \citep{qian2020thinking}, GramNet \citep{liu2020global}, PatchFron \citep{chai2020makes}, Swin-T \citep{liu2021swin}, Patch5M \citep{ju2022fusing}, SIA \citep{sun2022information}, LNP \citep{liu2022detecting} and LGrad \citep{tan2023learning}. Swin-T is a backbone network based on the Transformer \citep{Vaswani2017attention} and other methods are designed for forgery detection. We use the official codes and configurations to train and evaluate these methods.

\subsubsection{Cross-domain Evaluations}
For \textbf{cross-generator} evaluation, the methods are evaluated independently on GenImage and DF. For GenImage, following \citep{zhu2023genimage}, the methods are trained on SD v1.4 \citep{rombach2022high} training subset and evaluated on other evaluating subsets. For DF, following \citep{wang2023dire}, the methods are trained on real images and ADM-generated images \citep{dhariwal2021diffusion}, and evaluated on the evaluating sets. For \textbf{cross-scene} evaluation, the methods are trained on SD v1.4 \citep{rombach2022high}, ADM \citep{dhariwal2021diffusion} and Midjourney \citep{midjourney} subsets from GenImage (ImageNet scene). Then they are evaluated on the evaluating set generated by the corresponding generators from DF (LSUN-Bedroom scene).

\subsection{Model Training}
\subsubsection{Model Configurations}
For APN, Swin-Transformer-tiny \citep{liu2021swin} is used as the backbone in the spatial branch and ResNet-18 \citep{he2016deep} is used as the backbones in the frequency branch. The dimension $C$ of artifact-related and artifact-irrelevant features is set to 256. The numbers of vector bases in $\Omega^r$ and $\Omega^{ir}$ are both set to 128. The number of proposals $K$ is set to 15 for GenImage and 5 for DF. The number of heads in the self-attention layer is 8. For \textbf{$q_\theta$}, two fully connected nerual networks are used to predict the variational mean and variance values of features. 

\subsubsection{Training Configurations}
An alternative training strategy is used to train APN and $q_\theta$. Every time APN is trained, $q_\theta$ is trained three times. $q_\theta$ is frozen when APN is trained and APN is frozen when $q_\theta$ is trained. Dataloaders, optimizers and schedulers are set separately for them. For APN, the batch size is 32. Adam optimizer is used to optimize APN with the learning rate set to 1e-4 except that the learning rate for the learnable vector bases $\Omega^r$ and $\Omega^{ir}$ is set to 1e-7. The learning rates decay to 0.1 times its original values after every 20,000 iterations. For $q_\theta$, 32 different images are input into it for a batch. The learning rate of Adam optimizer is set to 1e-4 and it decays to 0.1 times its original values after every 60,000 iterations. The parameters of them are initialized randomly while $\Omega^r$ and $\Omega^{ir}$ are initialized to $\left[ I, Z \right]$ and $\left[ Z, I \right]$ respectively, where $I$ is an identity matrix and $Z$ is a zero matrix. The number of epochs is set to 10 for GenImage and 40 for DF for both networks. $\epsilon$ in Eq.\ref{eq:band} is set to 0.01. The images for both networks are randomly cropped to $224*224$ in the training stage and centrally cropped to $224*224$ in the inference stage. No other data augmentation methods are used.

\subsection{Cross-generator Evaluation}
\label{sec:cross-generator}
We first evaluate the cross-generator detection performance on GenImage. The results are shown in Tab.\ref{tab:cross-generator}. For most methods, the detection accuracy on SD v1.4, SD v1.5 and Wukong is consistently high. However, the accuracy on other generators drops significantly. For example, F3Net achieves the best performance on SD v1.4, SD v1.5 and Wukong, but only approaches random guessing accuracy on the other generators. It results in an average accuracy of only 68.7\% on GenImage. LNP achieves the best performance on Midjourney-generated images and the second performance on GLIDE-generated images while it lags behind other methods on other generators. And the same situation applies to LGrad. On the contrary, our proposed APN achieves best performance on 4 out of 8 generators and second on 3. Especially, for the generator GLIDE, the accuracy of APN is $21.0\%$ higher than the best value among other methods. As a result, the average detection accuracy of our method achieves $80.4\%$, which is at least $5.6\%$  higher than others. APN has consistent generalization superiority across various generators.

Tab.\ref{tab:DF} reports the cross-generator results on DF.  Overall, the difficulty of cross-generator detection on DF is lower than GenImage. F3Net and Patch5M achieve 96.8\% and 97.0\% average detection accuracy respectively. When we focus on our proposed method, it can be seen that APN achieves the best performance on 6 generators and the second on the rest generators. The average performance of APN on DF is 98.7\%, which is 1.7\% higher than the best value among other methods. In summary, APN has the better cross-generator detection capability compared with previous methods.

\begin{table*}[t]
    \scriptsize
    \renewcommand\arraystretch{1.5}
    \centering
    \caption{The cross-generator results on GenImage \citep{zhu2023genimage} (ImageNet scene \citep{deng2009imagenet}). The generators include SD v1 \citep{rombach2022high}, Wukong \citep{wukong}, ADM \citep{dhariwal2021diffusion}, BigGAN \citep{brock2018biggan}, GLIDE \citep{nichol2022glide}, Midjourney \citep{midjourney} and VQDM \citep{gu2022vector}. All methods are trained on the SD v1.4 subset. Accuracy ($\%$) is reported. \textbf{Bold} represents the best and \uline{underline} the second.}
    \begin{tabular}{cc | cccccccc|c}
    \toprule
    \multirow{1}{*}{Method} & Publication & \tabincell{c}{SD v1.4 \\} & \tabincell{c}{SD v1.5 \\} & \tabincell{c}{Wukong } & \tabincell{c}{ ADM } & \tabincell{c}{ BigGAN } & \tabincell{c}{ GLIDE} & \tabincell{c}{ Midjourney } & \tabincell{c}{VQDM } & Average \\
    \midrule
    Spec \citep{zhang2019detecting} & WIFS & 99.4 & 92.9 & 94.8 & 49.7 & 49.8 & 49.8 & 52.0 & 55.6 & 68.8 \\
    CNNSpot \citep{wang2020cnn}& CVPR & 96.3 & 95.9 & 78.6 & 50.1 & 46.8 & 39.8 & 52.8 & 53.4 & 64.2 \\
    F3Net \citep{qian2020thinking} & ECCV &\textbf{99.9}  & \textbf{99.9} & \textbf{99.9} & 49.9 & 49.9 & 50.0 & 50.1 & 49.9 & 68.7\\
    GramNet \citep{liu2020global} & CVPR  &99.2  & 99.1 & 98.9 & 50.3 & 51.7 & 54.6 & 54.2 & 50.8 & 69.9\\
    PatchFron \citep{chai2020makes} & ECCV & 99.7  & 99.4 & 96.7 & 51.0 & 51.6 & 54.1 & 66.2 & 54.1 & 71.6 \\
    Swin-T \citep{liu2021swin} & ICCV & \textbf{99.9} & \uline{99.8} & 99.1 & 49.8 & 57.6 & 67.6 & 62.1 & \uline{62.3} & \uline{74.8}\\
    Patch5M \citep{ju2022fusing} & ICIP  & \textbf{99.9} & \textbf{99.9} & \uline{99.8} & 50.6 & 50.4 & 51.7 & 71.6 & 51.0 & 71.9 \\
    
    SIA \citep{sun2022information} & ECCV  & \uline{99.8}  & 99.6 & 98.0 & 52.0 & 50.3 & 52.3 & 71.6 & 51.4 & 71.9\\
    LNP \citep{liu2022detecting} & ECCV & 94.8 & 94.9 & 88.7 & 50.8 & 56.5 & \uline{68.3} & \textbf{73.3} & 57.4 & 73.1 \\
    LGrad \citep{tan2023learning} & CVPR & 90.3 & 64.7 & 63.9 & \uline{52.2} & \textbf{69.9} & 62.7 & 56.5 & 52.0 & 64.0 \\
    APN (Ours) & - & \textbf{99.9} & \uline{99.8} & 98.9 & \textbf{53.9} & \uline{61.7} & \textbf{89.3} & \uline{72.3} & \textbf{67.2} & \textbf{80.4} \\
    \bottomrule
    \end{tabular}
    \label{tab:cross-generator}
\end{table*}

\begin{table*}[t]
\renewcommand\arraystretch{1.5}
    \scriptsize
    \centering
    \caption{The cross-generator results on DF \citep{wang2023dire} (LSUN-Bedroom scene \citep{yu2015lsun}). The generators include ADM \citep{dhariwal2021diffusion}, DALLE2 \citep{Ramesh2022HierarchicalTI}, Diff-ProjectedGAN (D-Pro.GAN) \citep{wang2022diffusion}, iDDPM \citep{nichol2021improved}, IF \citep{saharia2022photorealistic}, Midjourney (Mid.) \citep{midjourney}, ProjectedGAN (Pro.GAN) \citep{sauer2021projected} and SD v1 \citep{rombach2022high}. All methods are trained on ADM-generated images. Accuracy ($\%$) is reported. \textbf{Bold} represents the best and \uline{underline} the second.}
    \begin{tabular}{ cc | ccccccccc | c}
    \toprule
    Method & Publication & Real & ADM & DALLE2 & D-Pro.GAN & iDDPM & IF & Mid. &  Pro.GAN & SD v1 & Average \\
    \midrule
    \tabincell{c}{Spec \citep{zhang2019detecting} }  &  WIFS & 76.5 & 83.3 & 98.4 & 20.6 & 76.1 & 36.4 & 0.0 & 20.7 & 20.5 & 48.6\\
    \tabincell{c}{CNNSpot  \citep{wang2020cnn}}& CVPR & 38.4 & 70.0 & 85.4 & 58.9 & 65.4 & 82.4 & 61.0 & 62.4 & 78.5 & 66.9\\
    \tabincell{c}{F3Net  \citep{qian2020thinking}} & ECCV & 96.9 & \textbf{99.9} & \uline{99.8} & 86.9 & \uline{99.9} & \uline{99.9} & \textbf{100.0} & 87.9 & \textbf{99.9} & 96.8\\
    \tabincell{c}{GramNet \citep{liu2020global}} & CVPR & 75.5 & 79.9 &22.8&39.1&87.1&76.6&35.0&36.3&39.6&54.7\\
    \tabincell{c}{PatchFron  \citep{chai2020makes}} & ECCV & 77.6&72.8&26.4&32.9&78.1&57.4&28.0&34.1&46.2&50.4 \\
    \tabincell{c}{Swin-T \citep{liu2021swin}} & ICCV &71.9&71.0&55.8&39.0&78.3&39.5&56.0&43.6&71.7&58.5 \\
    \tabincell{c}{Patch5M \citep{ju2022fusing}} & ICIP &\textbf{99.8} &98.8&97.0&\uline{91.2} &98.5&\uline{99.9} & 97.0&\uline{91.8} & 99.1&\uline{97.0} \\
    \tabincell{c}{SIA  \citep{sun2022information}} & ECCV & \uline{99.5} &99.3&61.2&36.1&99.4&\uline{99.9} & 84.0&52.2&92.1&80.4\\
    LNP \citep{liu2022detecting} & ECCV & 93.7 & 93.9& 94.2 & 28.5 & 89.9 & \uline{99.9} & 91.0 & 31.6 & 69.8 & 76.9\\
    LGrad \citep{tan2023learning} & CVPR  & 97.9 & 98.9 & 90.2 & 67.0 & 97.1 & 99.5 & 70.0 & 68.3 & 98.9 & 87.5\\
    APN (Ours) & - & \textbf{99.8} &\uline{99.6} & \textbf{100.0}&\textbf{94.3}&\textbf{100.0}&\textbf{100.0}&\uline{99.0}&\textbf{96.0}&\uline{99.3} & \textbf{98.7}  \\
    \bottomrule
    \end{tabular}
    \label{tab:DF}
\end{table*}

\begin{table*}[t]
\renewcommand\arraystretch{1.5}
    \scriptsize
    \centering
    \caption{The cross-scene results on three generators, i.e. SD v1 \citep{rombach2022high}, ADM \citep{dhariwal2021diffusion}, and Midjourney  \citep{midjourney}. All methods are trained on ImageNet scene and evaluated on ImageNet and LSUN-Bedroom (LSUN) scene. Accuracy ($\%$) is reported. \textbf{Bold} represents the best and \uline{underline} the second.}
    \begin{tabular}{cc | cc  | cc  | cc  | ccc }
    \toprule
    \multirow{2}{*}{Method} & \multirow{2}{*}{Publication} & \multicolumn{2}{c|}{\makecell[c]{SD v1 }} &  \multicolumn{2}{c|}{ \makecell[c]{ADM}}  &  \multicolumn{2}{c|}{ \makecell[c]{Midjourney}} & \multicolumn{3}{c}{Average}  \\
     & &  ImageNet   & LSUN &  ImageNet   & LSUN &  ImageNet   & LSUN & ImageNet & LSUN & Both\\
    \midrule
    Spec \citep{zhang2019detecting} & WIFS & 99.4 & 96.5 & 97.3 & 53.0  & 98.0 &  93.5 &98.2&81.0&89.6 \\
    CNNSpot \citep{wang2020cnn} & CVPR & 96.3  &  97.2 & 99.1 &\uline{90.2} & \uline{99.7} & 98.4 & 98.4 & \uline{95.3} & 96.8\\
    F3Net \citep{qian2020thinking} & ECCV  & \textbf{99.9} & 97.3 & \uline{99.5} & 89.2 & \uline{99.7} & \uline{99.0} & \textbf{99.7} & 95.2&\uline{97.4}\\
    GramNet \citep{liu2020global} & CVPR & 99.2 & 91.6& 94.8 & 72.3 & 94.0 &88.2 & 96.0&84.0&90.0\\
    PatchFron \citep{chai2020makes} & ECCV & 99.7 & 97.1  & 97.6  &  78.1 & 97.9  & 96.6  & 98.4&90.6&94.5\\
    Swin-T \citep{liu2021swin} & ICCV & \textbf{99.9}  & 86.2  & 93.3 & 70.1  &  93.6 & 90.0  &95.6&82.1& 88.9\\
    Patch5M \citep{ju2022fusing} & ICIP & \textbf{99.9} & \uline{97.5}  & 99.4 & 79.1 & \textbf{99.8} &  97.3&\textbf{99.7} & 91.3&95.5\\
    
    SIA \citep{sun2022information} & ECCV & \uline{99.8}  & 97.0  & 99.0  & 88.3  & 99.6 & 98.9 & 99.5&94.7&97.1\\
    LNP \citep{liu2022detecting} & ECCV & 94.8 & 91.1 & 94.3 & 59.2 & 90.2 & 84.9 & 93.1 & 78.4 & 85.8  \\
    LGrad \citep{tan2023learning} & CVPR & 90.3 & 84.5 & 85.6 & 54.7 & 78.0 & 85.1 & 84.6 & 74.8 & 79.7\\
    APN (Ours) & - & \textbf{99.9} &  \textbf{100.0} &  \textbf{99.9} &  \textbf{99.2} &  99.1 & \textbf{99.2}  &  \uline{99.6}&\textbf{99.5}&\textbf{99.6} \\
    \bottomrule
    \end{tabular}
    \label{tab:cross-scenario}
\end{table*}

\subsection{Cross-scene Evaluation}
\label{sec:cross-scenario}

\begin{figure*}[t]
  \centering
   \includegraphics[scale=0.23]{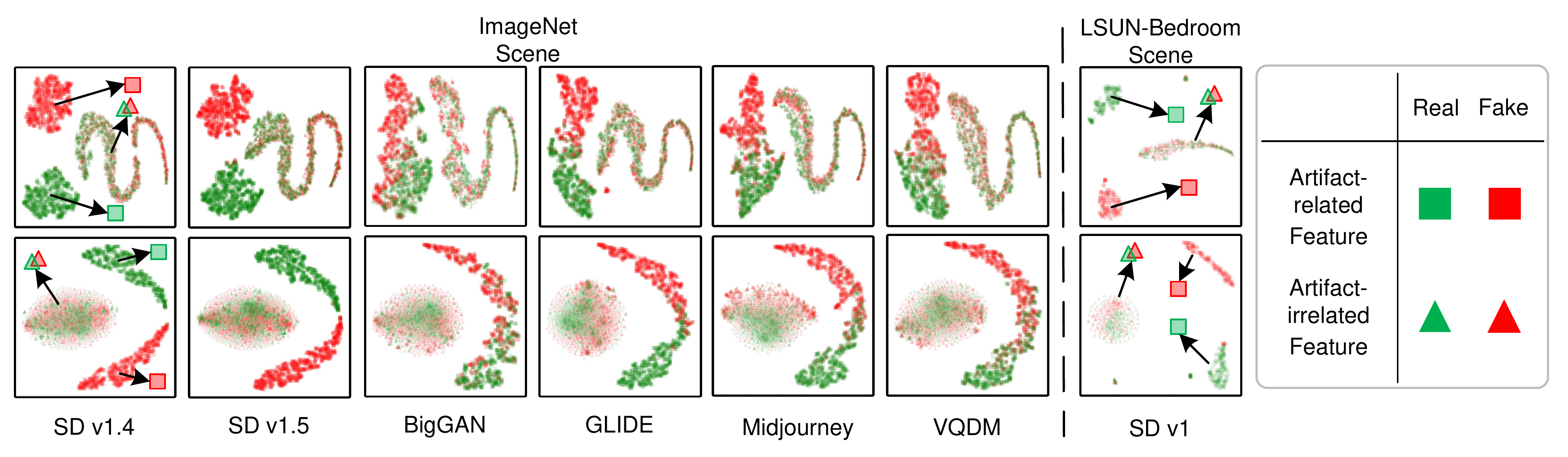}
   \caption{The visualization of artifact-related and -irrelated features of samples in different scene and generator domains with / without the implicit purification using t-SNE. The top row: APN with the implicit purification. The bottom row: APN without the implicit purification. Please zoom in to see the details.}
   \label{fig:feature}
\end{figure*}

 Tab.\ref{tab:cross-scenario} shows the cross-scene evaluation results of previous methods and ours. For LSUN-Bedroom scene, we report the average detection accuracy for real images and corresponding generator-generated images because the subsets for ImageNet scene from GenImage include real images. The results demonstrate that when the previous methods are faced with cross-scene images, the detection accuracy will decrease to varying degrees even if they are generated by same generators. CNNSpot and F3Net are the ones with the best cross-scene detection performance among previous methods. Nonetheless, their average detection accuracy on LSUN-Bedroom scene drops 3.1\% and 4.5\% respectively compared with the corresponding accuracy on ImageNet scene. On the contrary, our method maintains a high detection performance across scenes. The average accuracy on LSUN-Bedroom scene is only 0.1\% lower than the one on ImageNet scene. Compared with other methods, APN achieves the best performance in all three experiments on LSUN-Bedroom scene. In summary, APN achieves a better cross-scene detection performance compared with previous methods.

\subsection{Visualizations}
\label{sec:vis}

\begin{figure}[t]
  \centering
   \includegraphics[scale=0.3]{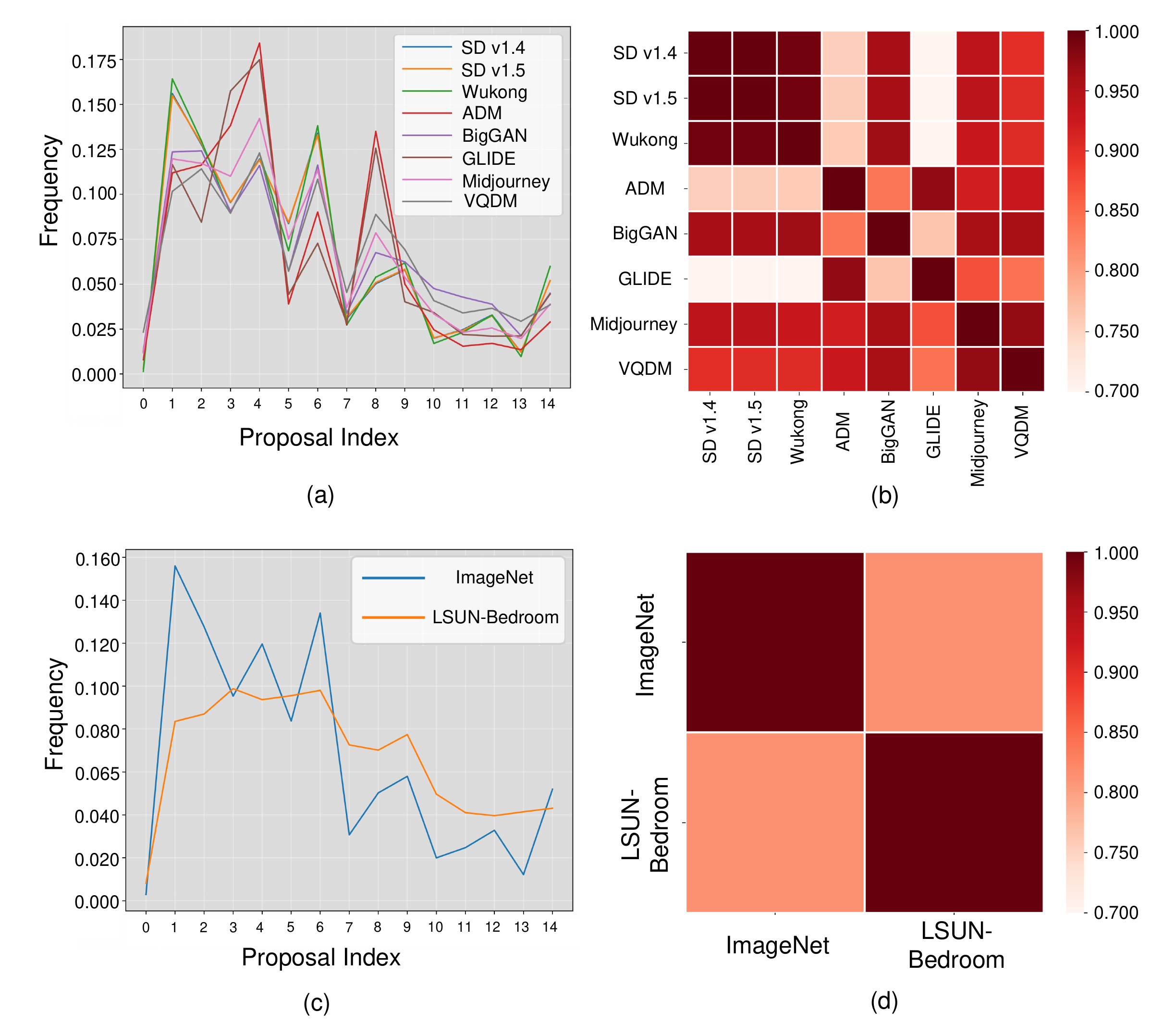}
   \caption{The differences in proposals across generators and scenes. (a) and (c): the confidence-weighted and normalized frequencies of the proposals. (b) and (d): the correlation matrices of the curves in (a) and (c). (a) and (b) correspond to the fake images in GenImage subsets. (c) and (d)  correspond to the fake images generated by SD v1.4 in ImageNet (GenImage) and LSUN-Bedroom (DF) scene.}
   \label{fig:freq_curve}
\end{figure}

\begin{figure}[h]
  \centering
   \includegraphics[scale=0.16]{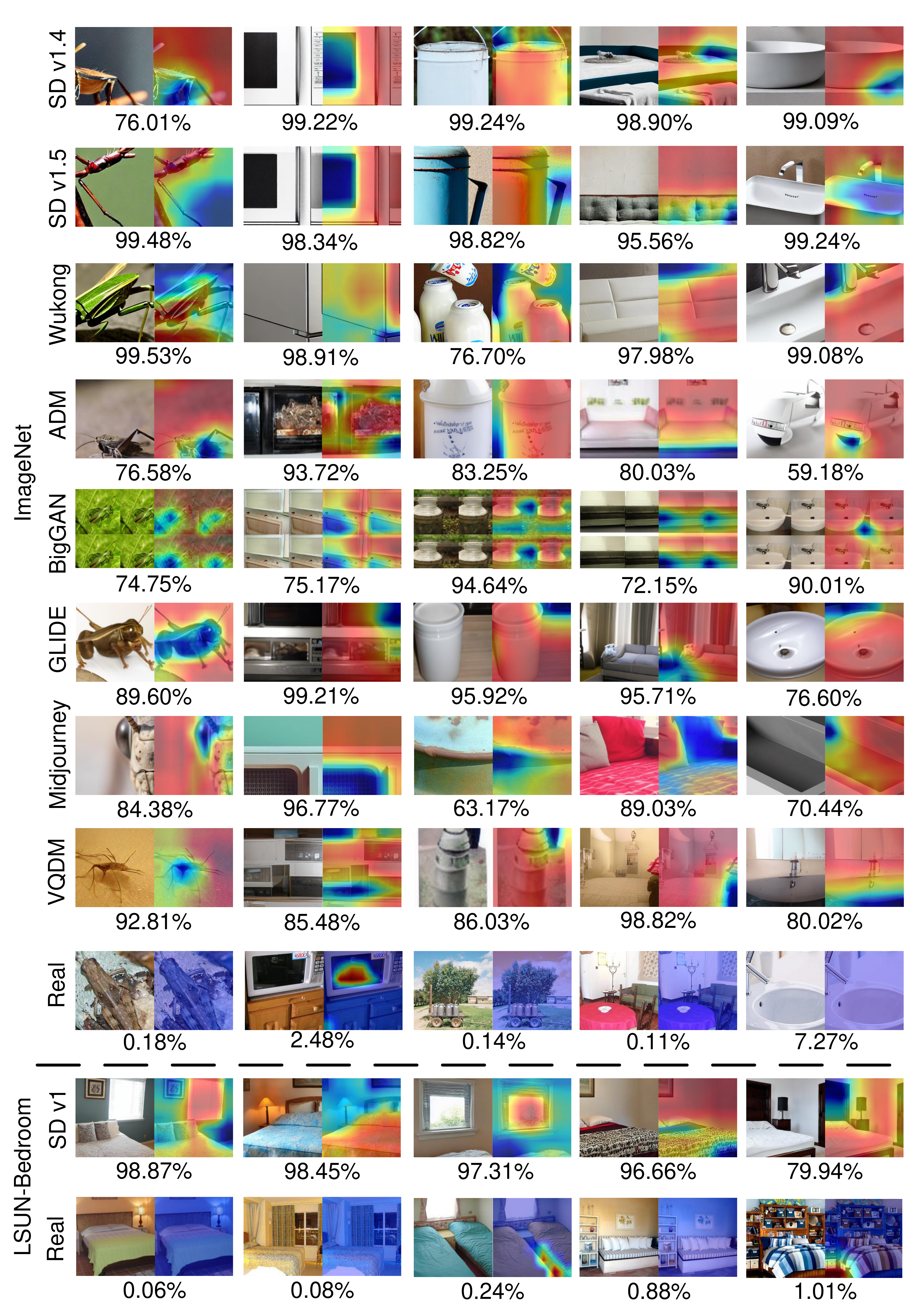}
   \caption{The images and their spatial heatmaps of GenImage (ImageNet scene) and DF (LSUN-Bedroom scene) samples. The predicted probabilities are below the images. The images in each column in ImageNet scene have the same class label.}
   \label{fig:spatial}
\end{figure}

\begin{figure}[ht]
  \centering
   \includegraphics[scale=0.3]{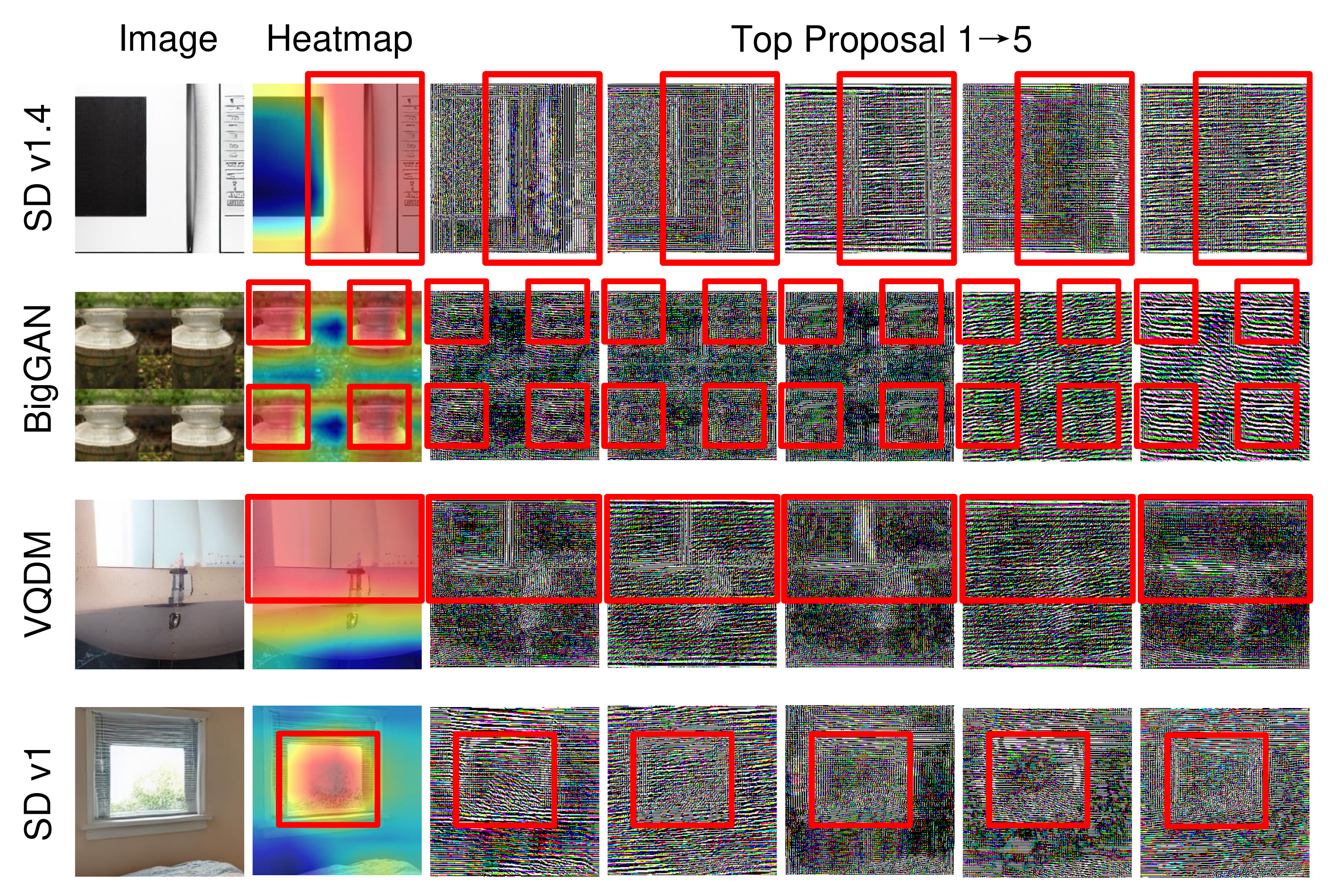}
   \caption{The spatial images of the top-5 frequency proposals. The samples are from Fig.\ref{fig:spatial}. It shows the local (e.g. the 2nd proposal in the 1st row and the 3rd proposal in the 4th row) and global (e.g. the 4th proposal in the 2nd row and the 4th proposal in the 3rd row) noise pattern examples. The local noise patterns often exist in the warm or hot areas in the corresponding heatmaps.}
   \label{fig:proposal}
\end{figure}

With the above results, we want to further justify the following two questions: \emph{Why can APN improve cross-domain detection performance? And what are the cross-domain commonalities of artifact features in space that APN focuses on?} In this subsection, we attempt to answer them through visualizations of artifact-related and -irrelated features, frequency proposals and spatial heatmaps. All visualizations are based on the model trained on SD v1.4 in GenImage.

\textbf{Why can APN improve cross-domain detection performance?} We attempt to answer it from the view of two key components of APN, i.e. the implicit purification and the frequency-band proposal method. For the implicit purification, we visualize the artifact-related and -irrelated features in GenImage and DF with and without the implicit purification in Fig.\ref{fig:feature}. From the figure, we can find two main differences between the features with and without the implicit purification. On the one hand, without the implicit purification, the distribution of artifact-irrelated features is different between real and fake samples. It is manifested as a difference in the distribution density of the two features at the same location in the figure. This truth means that the artifact-irrelated features still contain artifact information. On the contrary, the difference in the distribution is smaller with the implicit purification. The artifact-irrelated features for real and fake images are more consistently distributed in the space. On the other hand, with the implicit purification, the distribution of the artifact-related features is closer to an isotropic distribution, which means a higher intra-class similarity. It implies that the implicit purification enhances the consistency of artifact features across different images, thereby improving the cross-domain performance of APN. In summary, the implicit purification condenses the forgery information implicit in irrelevant features, and further enhances the consistency of artifact features in various images.

For the frequency-band proposal method, we find that given the index of a proposal, the frequency range covered by the proposal is close across images. It may be related to the fact that we add a prior position for each proposal and APN is only trained on one generator. Then we count the confidence-weighted proposal frequencies among subsets and plot them as the curves in Fig.\ref{fig:freq_curve}. The correlation matrices among the curves are also calculated. It can be seen that statistically speaking, the proposals have different importance for different generators and scenes. It indicates that the proposals around the same frequency play different roles in forgery detection across generators and scenes. We also notice that the correlations among SD v1.4, SD v1.5 and Wukong are high. It is consistent with the fact shown in Fig.\ref{tab:cross-generator} that the performance of the methods on the three generators is highly consistent. It demonstrates that the frequency artifacts of the three generators are very similar. In summary, the proposal method adds flexibility in frequency artifact extraction and reduces the risk of overfitting to a fixed pattern in the training set. 


\textbf{What are the cross-domain commonalities of artifact features in space that APN focuses on?} We attempt to answer it through spatial heatmaps given by Layer-CAM \citep{jiang2021layercam}. The target layer is set to the final layer of the spatial backbone. The results are shown in Fig.\ref{fig:spatial}. We have two observations about the results. First, the area of the activation regions is relatively large for fake images while there is small or even no activation area for real images. It shows that APN extracts global artifact features in the spatial modality. Second, the regions that APN focuses on in the images exhibit diversity. We can find that it focuses on bright (e.g. row 3 and column 4, and row 4 and column 5) or textured (e.g. row 7 and column 4, and row 10 and column 2) areas while sometimes it also focuses on dark (e.g. row 1 and column 1, row 2 and column 5) or plain (e.g. row 5 and column 4, row 6 and column 5) areas. The finding is consistent across both domains. It shows that our method learns diverse spatial artifact features from a single training set. In summary, the commonalities of artifact features in space that APN focuses on across domains are globality and diversity.

Finally, we further investigate the relationships between heatmaps and frequency proposals. The spatial images of the top-5 proposals with the highest confidence scores are plotted with their corresponding heatmaps in Fig.\ref{fig:proposal}. We find that the content of these images is high-frequency noise and cannot be understood by humans. Some noise patterns in proposals are global while others often exist in the hot or warm areas in the heatmaps. For example, for the 2nd proposal in the 1st row in Fig.\ref{fig:proposal}, the noise texture in the corresponding warm-toned area in the heatmap is significantly different from that in other areas. And for the 3rd proposal in the 4th row in Fig.\ref{fig:proposal}, the noise intensity in the corresponding warm-toned area in the heatmap is stronger than other areas. It shows some correlation between spatial and frequency modality in the extracted forgery clues.

\begin{table}[ht]
    \centering
    \caption{The ablation results. (a) The module ablation results on GenImage. EP-F: Explicit Purification in Frequency; PPS: the proposal method; EP-S: Explicit Purification in Space; D: the spatial feature decomposition method; IP: Implicit Purification. \checkmark denotes the model with the corresponding module. For ease of reading, the line number of each line (Line) is given. (b) The proposal number ablation results on GenImage and DF. The average accuracy (\%) on all subsets is reported.}
    \begin{minipage}[t]{0.55\textwidth}
	\subfloat[]{
        \begin{tabular}{c|cc|cc|c | c }
        \toprule
        \multirow{2}{*}{Line} & \multicolumn{5}{c|}{Ablation module } & \multirow{2}{*}{Avg.}\\ 
        &EP-F & PPS & EP-S & D & IP &   \\
        \midrule
        1 & \checkmark & \checkmark & \checkmark & \checkmark &  & 75.3 \\
        2 & & & & & \checkmark &  71.0\\ \midrule
        3&\checkmark & \checkmark & & & \checkmark  & 79.1\\
        4& & & \checkmark & \checkmark & \checkmark  & 77.1\\ \midrule
        5&\checkmark & & \checkmark & \checkmark & \checkmark  & 74.8\\
        6&\checkmark & \checkmark & \checkmark & & \checkmark  & 77.8\\ \midrule
        7&\checkmark & \checkmark & \checkmark & \checkmark & \checkmark & \textbf{80.4}\\
        \bottomrule
        \end{tabular}} 
    \end{minipage}
    \begin{minipage}[t]{0.3\textwidth}
    \subfloat[]{
        \renewcommand{\arraystretch}{1.49}
        \begin{tabular}{c | c c }
        \toprule
        \multirow{1}{*}{\# of proposals} & GenImage & DF \\
        \midrule
        1 & 72.9 & 89.0 \\
        5 & 72.6 & \textbf{98.7}\\
        10 & 75.2 &  \textbf{98.7}\\
        15 & \textbf{80.4}  & 98.2\\
        20 & 77.3  & 98.6\\
        25 &  76.5 & 98.6\\
        \bottomrule
        \end{tabular}}
    \end{minipage}
    \label{tab:abalition}
\end{table}

\subsection{Ablation Experiments}
\label{sec:ablation}
APN mainly includes three parts, i.e. the explicit purification in frequency, the explicit purification in space and the implicit purification. And the suspicious frequency-band proposal method and the spatial feature decomposition method are proposed for the explicit purification. To show the roles of these modules, the ablation experiments are conducted in this subsection. The experiments follow the setting on GenImage and the average accuracy on all subsets is reported. The results are shown in Tab.\ref{tab:abalition}(a). To show the role of the implicit purification (Line 1 in Tab.\ref{tab:abalition}(a)), we train APN without the mutual information estimation, i.e. remove $\mathcal{L}_{mi}$ from the loss function $\mathcal{L}$ in Eq.\ref{eq:loss}. To show the role of the explicit purification (Line 2), a Swin-Transformer-tiny is used to extract features from images directly, and the features are split into the artifact-related features and the artifact-irrelated features along the channel dimension. The implicit purification is implemented on them. To show the roles of the space and frequency modality (Line 3 and 4), the two branches and the corresponding loss functions are removed respectively. The implicit purification is conducted on the features of one branch directly. For ablation of the proposal method (Line 5), a ResNet-50 backbone is used to extract the artifact-related features from the spectra of inputs. For ablation of the spatial feature decomposition method (Line 6), it is replaced with splitting along the channel dimension.  For each ablation experiment, the unmentioned parts of APN remain unchanged. 

Tab.\ref{tab:abalition}(a) mainly reflects four facts. First, the implicit purification improves the cross-domain detection by $5.1\%$ (from $75.3\%$ in Line 1 to $80.4\%$ in Line 7) while the explicit purification provides more vital artifact features (from $71.0\%$ in Line 2 to $80.4\%$ in Line 7). Second, compared with the spatial modality (Line 3), more performance gain is brought by the frequency modality (Line 4, $2.0\%$ higher). Third, the proposal and decomposition methods both significantly improve performance (from $74.8\%$ in Line 5 to $80.4\%$ in Line 7 for the proposal method, and from $77.8\%$ in Line 6 to $80.4\%$ in Line 7 for the spatial decomposition method). Finally, the introduction of one modality without the corresponding separation method hurts the performance (drops from $77.1\%$ in Line 4 to $74.8\%$ in Line 5 for frequency, and from $79.1\%$ in Line 3 to $77.8\%$ in Line 6 for space modality). 

Tab.\ref{tab:abalition}(b) shows the results on GenImage and DF using different numbers of proposals. It shows that the optimal proposal number differs for different datasets. A difficult dataset, such as GenImage, requires more proposals to cope with more artifacts.

\section{Conclusion}
In this paper, we propose APN to improve cross-generator and cross-scene generated image detection. It purifies domain-agnostic artifact features by 1) the explicit purification based on the frequency proposals and the spatial feature decomposition and by 2) the implicit purification based on the mutual information estimation. Experiments show that compared with previous methods, APN achieves better results across generators and scenes on two datasets. We also discuss the effects of the implicit purification and proposals, and the commonalities of the features in space that APN focuses on across domains. APN provides an effective tool to identify generated images in the era of AIGC.